\documentclass{article}

\usepackage{microtype}
\usepackage{graphicx}
\usepackage{subfigure}
\usepackage{booktabs} 
\usepackage{bm}
\usepackage{amsmath}
\usepackage{amsfonts}
\usepackage{xcolor}
\usepackage{comment}

\usepackage{hyperref}

\newtheorem{theorem}{Theorem}
\newtheorem{definition}{Definition}
\newtheorem{prop}{Proposition}
\def\diag{\mathop{\rm diag}\nolimits}
\DeclareMathOperator{\Tr}{Tr}

\usepackage[accepted]{icml2021}

\icmltitlerunning{Minimum sharpness: Scale-invariant parameter-robustness of neural networks}

\begin{document}

\twocolumn[
\icmltitle{Minimum sharpness: Scale-invariant parameter-robustness of neural networks}




\begin{icmlauthorlist}
\icmlauthor{Hikaru Ibayashi}{usc}
\icmlauthor{Takuo Hamaguchi}{ut}
\icmlauthor{Masaaki Imaizumi}{ut}
\end{icmlauthorlist}

\icmlaffiliation{usc}{Department of Computer Science, University of Southern California, Los Angeles, CA, USA}
\icmlaffiliation{ut}{Komaba Institute for Science,  University of Tokyo, Tokyo, Japan}

\icmlcorrespondingauthor{Hikaru Ibayashi}{ibayashi@usc.edu}
\icmlkeywords{Machine Learning, Deep Learning, Model Robustness, Generalization of Neural Network}

\vskip 0.3in
]



\printAffiliationsAndNotice{}  

\begin{abstract}
Toward achieving robust and defensive neural networks, the robustness against the weight parameters perturbations, i.e., sharpness, attracts attention in recent years (Sun et al., 2020).
However, sharpness is known to remain a critical issue, ``scale-sensitivity.”
In this paper, we propose a novel sharpness measure, \textit{Minimum Sharpness}.
It is known that NNs have a specific scale transformation that constitutes equivalent classes where functional properties are completely identical, and at the same time, their sharpness could change unlimitedly.
We define our sharpness through a minimization problem over the equivalent NNs being invariant to the scale transformation.
We also develop an efficient and exact technique to make the sharpness tractable, which reduces the heavy computational costs involved with Hessian.
In the experiment,
we observed that our sharpness has a valid correlation with the generalization of NNs and runs with less computational cost than existing sharpness measures.
\end{abstract}
\section{Introduction}
Despite the tremendous success of neural networks (NNs), NNs are known to be vulnerable to adversarial examples, i.e., simple perturbations to input data can mislead models \cite{Goodfellow2014-ct, kurakin2016adversarial, Wu2017-ox}.
Many research attempts to resolve this vulnerability to make NNs defensive for the noisy input data.
Similarly, the robustness against the model parameters perturbation, or parameter-robustness, is equally essential to make NNs defensive against the noise coming from hardware neural networks \cite{lecun20191}.

The parameter-robustness is a well-studied topic in different lines of research, where it is called ``sharpness" \cite{Hochreiter1997-xr, Keskar2016-tn}.
Although sharpness was initially studied in connection to generalization performance, an increasing number of results show that controlling sharpness is the effective way to design robust models.
For example, \citet{Foret2020-kh} proposed  ``Sharpness-Aware Minimization" that minimizes approximated sharpness measure in their training process. Similarly, \citet{Sun2020-zw} proposed ``adversarial corruption-resistant training” that implements adversarial corruption of model parameters in their training process.
As a theoretical justification, \citet{wei2019improved} rigorously proved that sharpness regarding the activation functions gives a tight upper bound of the NNs performance.

However, despite its effectiveness being shown in many ways, sharpness leaves a critical unsolved problem; ``scale-sensitivity."
\citet{pmlr-v70-dinh17b} has pointed out that traditional sharpness measures are problematic under scale transformation.
If NNs utilize a non-negative homogeneous function, $\phi(a x) = a \phi(x)$ for all $a > 0$ such as ReLU \cite{ICML-2010-NairH} and Maxout~\cite{pmlr-v28-goodfellow13} activation function, then a certain scale transformation on parameters does not change the functional property at all;
however, the naive sharpness measures may change significantly.
This unexpected behavior poses a problem that good sharpness should be invariant under such scale transformation, but naive sharpness measures do not.

The following are the previous major studies tackling the scale-sensitivity problem.
The first study \cite{rangamani2019scale} proposed sharpness measure as a spectral norm of Riemannian Hessian on a quotient manifold. 
The quotient manifold makes the sharpness measure invariant due to the definition over equivalence relation via $\alpha$-scale transformation.
The second study \cite{icml2020_3399} developed a sharpness measure through the minimization problem.
The minimization enables us to select sharpness independent of scaling parameters.
Although those proposed sharpness measures are free from scale-sensitivity, they have the following drawbacks: (i) they require intractable assumptions in their derivation process, and (ii) they suffer from heavy computation to handle Hessian matrices.

In this paper, we propose a novel invariant sharpness measure called \textit{Minimum Sharpness}, which overcomes the aforementioned problems.
Our sharpness measure is defined as a minimum trace of Hessian matrices over equivalence classes generated by scale transformation.
We show that our sharpness measure is scale-invariant owing to the minimization over an equivalence class.
Further, our measure is computationally efficient thanks to our technique that can exactly and efficiently calculate Hessian.
With this technique, the minimization can be carried out without costly computations, but with just several epochs forward-and-back propagation.

As empirical justifications, we carried out an experiment to report the efficiency and accuracy of our technique compared to the ground-true Hessian trace calculation.
We also empirically confirmed that our sharpness validly correlates with the performance of models.
The implementation of the experiment is available online\footnote{\url{https://github.com/ibayashi-hikaru/minimum-sharpness}}.
    
\section{Methodology}
\label{algorithm}

\subsection{Preliminary}

\paragraph{Neural Network}:
We define fully-connected neural networks (FCNNs) as follows:
\begin{equation*}
  f(x\mid \theta) 
  = 
    \bm{W}_{D} \phi 
    \left(
      \bm{W}_{D-1} \phi 
      \cdots
      \left( \phi 
        \left( \bm{W}_{1} \mathbf{x} \right) \cdots
      \right)
    \right)
\end{equation*}
where
$\mathbf{x}$ is an input,
$D$ is the number of layers, 
$\theta = \{\bm{W}_{d}\}_{d=1}^D$ is a set of weight parameter matrices of $d$-th layer (layer parameter),
and $\phi$ is an activation function.
For brevity, we use only FCNNs in this study to the explanation.
\footnote{The same discussion is applicable with various NNs, including convolutional NNs.}

\paragraph{$\alpha$-Scale Transformation}:
$\alpha$-Scale transformation \cite{pmlr-v70-dinh17b} is a transformation of parameters of FCNNs that does not change the function.
We use $\alpha=\{\alpha_d\}_{d=1}^D$ such that 
  $\alpha_d > 0$ and $\prod_d \alpha_d=1$
  to rescale parameters of NNs as follows:
  $\theta' = \{\bm{W}_d' \}_{d=1}^D = \{\alpha_d\bm{W}_d\}_{d=1}^d$.
We denote the functionality of this transformation as $\theta' = \alpha(\theta)$.
Importantly, the scale transformation does not change a function by FCNNs, i.e. $\forall \mathbf{x}, f(\mathbf{x} \mid \theta ) = f(\mathbf{x} \mid \theta')$, if their activation functions $\phi$ are non-negative homogeneous (NNH) $\forall a>0, \phi(a x) = a \phi(x)$ such as Rectified Linear Unit (ReLU) activation.
\begin{figure*}[t]
	\begin{minipage}{100mm}
    \begin{tabular}{c|ccc}
  		\toprule
  		num-of-data     & $n=10$      & $n=100$      & $n=1000$    \\
  		\midrule
  		Baseline (FCNNs)  & $11.11$ sec & $11.05$ sec & $12.23$ sec \\
  		Proposal (FCNNs)  & $0.018$ sec & $0.019$ sec & $0.021$ sec \\
  		\midrule
  		Baseline (CNNs)   & $10.19$ sec & $10.73$ sec & $21.22$ sec \\
  		Proposal (CNNs)   & $0.026$ sec & $0.036$ sec & $0.334$ sec \\
  		\bottomrule
  	\end{tabular}
	\end{minipage}
	\begin{minipage}{50mm}
	\centering
    \includegraphics[width=\textwidth]{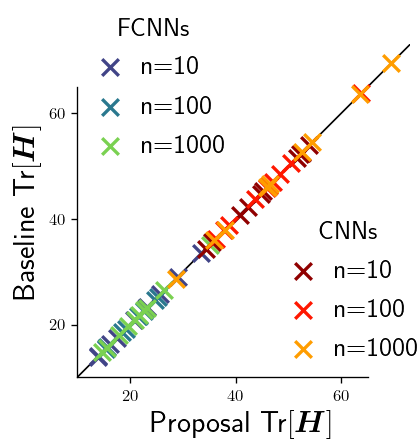}
	\end{minipage}
  \caption
  {
    Comparison between baseline and proposal for calculating $\Tr[\bm{H}]$
    .
    The left table reports averaged computational time, and the right figure shows the calculated trace. 
  }
  \label{res:prop:correct}
\end{figure*}
\subsection{Proposed Method: Minimum Sharpness}
The previous work \cite{pmlr-v70-dinh17b} pointed out that the $\alpha$-scale transformation can generate NNs whose functionalities are identical, but naive sharpness measures for NNs can be different.
Let $L(\theta) = L(f(\cdot | \theta))$ be a loss function with a function $f(\cdot | \theta)$ and $\bm{H}_{\theta} := \nabla_\theta^2 L(\theta)$ is a Hessian matrix, whose trace can be a proxy of sharpness.
Obviously $L(\theta) = L(\alpha(\theta))$ holds for any $\alpha$, but the trace of $\bm{H}_{\theta} $ can change under varying $\alpha$. 
This problem motivates us to develop the ``invariant'' sharpness measure to the $\alpha$-scale transformation.

\begin{definition}
\label{def:MS}
We propose the following novel sharpness named \textit{Minimum Sharpness} (MS) at $\theta$,
\begin{equation}
\label{eq:minimization}
  \text{MS}_{\theta}
  = 
  \min_{\alpha} 
    \Tr[ \bm{H}_{\alpha(\theta)}].
\end{equation}
\end{definition}

By its definition, the minimum sharpness is invariant to $\alpha$-scale transformation.

To enjoy the benefits of minimum sharpness,
  we need to overcome two difficulties.
The first is the high computational cost to calculate Hessian matrices
because a native computation of the whole Hessian matrix takes an infeasible amount of memory. 
The second difficulty, the target function of the optimization problem \eqref{eq:minimization} is unclear in a form with respect to $\alpha$. 
Otherwise, we have to solve the minimization problem using inefficient methods such as grid search.

To make minimum sharpness tractable, we exploit the following two propositions regarding the Hessian matrix.
In the following, we consider a classification problem with a softmax loss and $K$ labels.
We note that the same discussion is applicable to convolutional NNs and other loss functions.
The following results simplify the minimization problem \eqref{eq:minimization}.
\begin{prop}
\label{prop:BP}
Let 
  $\{(\mathbf{x},y)\}_i^n$ be a set of data pairs, 
  $o_l$ is logit for label $l$,
  $p_l = \frac{1}{Z} \exp[o_l]$,
  $Z=\sum_I \exp[o_l]$, 
  and 
  $L(\theta)= - \frac{1}{n} \sum_i \ln p(y_i \mid \mathbf{x}_i; \theta)$. 
Then we have following formulation 
\begin{eqnarray*}
  &&\Tr[\bm{H}_\theta] =
      \frac{1}{n}
      \sum_{i=1}^n
        \text{score}(\mathbf{x}_i,y_i), \nonumber\\ 
  &&\text{score}(\mathbf{x},y) 
  =
    \sum_{d=1}^D
      \left(
      \sum_{l = 1}^K
        p_l
        \left\|\frac{\partial o_l}{\partial \bm{W}_d}\right\|_F
    -
      \left\|\frac{\partial \ln Z}{\partial \bm{W}_d}\right\|_F
      \right) \nonumber\\ 
\end{eqnarray*}
where $\|\cdot\|_F$ denotes Frobenius norm.
\end{prop}
\begin{prop}
\label{prop:invrselaw}
Let $\bm{W}_d$ be a layer parameters in NNs with NNH activation function, $\alpha = \{\alpha_d\}_d$ be parameters of $\alpha$-scale transformation.
Then following relations hold
\begin{eqnarray*}
  \frac{\partial \ln Z}{\partial \bm{W}_d}
  = 
  \frac{1}{\alpha_d}
  \frac{\partial \ln Z'}{\partial \bm{W}_d'},
  \,\,\,\,\,\,\,\,\,\,
  \frac{\partial o_l}{\partial \bm{W}_d}
  = 
  \frac{1}{\alpha_d}
  \frac{\partial o_l'}{\partial \bm{W}_d'},
\end{eqnarray*}
where $\bm{W}_d' = \alpha_d \bm{W}_d$. 
The values $Z,o_l$ and $Z',o_l'$ are calculated using NNs $f$ and $\alpha(f)$ respectively.
\end{prop}
Both proofs are provided in Appendix~\ref{app:BP} and \ref{app:inverselaw} respectively.
The first one provides us with decomposition to calculate $\Tr[\bm{H}]$, exactly and efficiently.
That is, this calculation requires only $K+1$ epochs forward\&back-propagation for gradients of $Z$ and $o_l$. 
To the best of our knowledge, this decomposition is also our contribution.

Combining the two results, we have the following tractable formulation of minimum sharpness as
\begin{align*}
\text{MS}_\theta
  = 
  \min_{\alpha} 
    \sum_{d=1}^D \frac{1}{\alpha_d^2} 
      \Tr[\bm{H}_{\theta,d}],
\end{align*}
where
\begin{align*}
  \Tr[\bm{H}_{\theta,d}]
  =
    \frac{1}{n}
    \sum_{i=1}^n
      (
      \sum_{l=1}^K
        p_{l@i}
        \|\frac{\partial o_{l@i}}{\partial \bm{W}_d}\|_F
        -
      \|\frac{\partial \ln Z_{@i}}{\partial \bm{W}_d}\|_F), 
\end{align*}
and $\bm{H}_{\theta,d}$ denotes diagonal blocked Hessian over $d$-th layer and and subscription $@i$ indicates $i$-th data's outputs.
Here, applying inequality of arithmetic and geometric means
\begin{align*}
  \sum_d^D
    \frac{1}{\alpha_d^2} 
    \Tr[\mathbf{H}_{\theta,d}]
  &\geq 
  D 
  \prod_d 
    (
    \frac{1}{\alpha_d^2} 
    \Tr[\mathbf{H}_{\theta,d}]
    )^{1/D}& \\
  &= 
  D 
  \sqrt[D]
    {
    \prod_d 
    \Tr[\mathbf{H}_{\theta,d}]
    },&
\end{align*}
we obtain the following result.
We note that a more detailed explanation is available in Appendix~\ref{app:theorem}.

\begin{theorem}
\label{def:reform}
We can formulate minimum sharpness as follow  for no-bias FCNNs $f$ case:
\begin{align*}
  \text{MS}_\theta
    = 
    D
    \left(
      \prod_{d=1}^D 
    \Tr[\mathbf{H}_{\theta,d}]
    \right)^{1/D}.
\end{align*}
\end{theorem}

Our sharpness has the following remarkable benefits.
Firstly,
  we can check and interpret the invariance easily
  because of its simple design such that all scales $\alpha_d$ are canceled out.
Secondly,
  these results can be extended to with-bias NNs and convolutional NNs,
  e.g., by adding terms $({\prod_{i=1}^d \alpha_i})^{-2} \|\frac{\partial}{\partial \mathbf{b}_d}\|_2$ 
  for $d$-th layer's bias $\mathbf{b}_d$
  .
Thirdly,
  as shown in the experiment section,
  minimum sharpness correlated with generalization gaps 
  on the same level as previous invariant sharpness.
Lastly, owing to the form in Proposition \ref{prop:BP}, we can compute the measure very efficiently without errors.

\section{Experiments}
\label{experiments}
\subsection{Accuracy and Efficiency of  \texorpdfstring{$\textrm{Tr}[\bm{H}]$}{Tr[H]} Calculation}
\label{expe:eeTrH}
  We verify the accuracy and efficiency of our developed calculation of the trace of Hessian matrices.
  We compare the proposed method with a naive calculation (referred to as ``baseline'' here),
    which 
    first computes the gradients of all parameters,
    applies parameter-wise derivation naively for the gradients,
    and then selects diagonal elements from the outcome.\footnote{Note that we did NOT calculate Hessian of all parameter pairs. We tried several implementations within the naive formulation $\nabla \nabla L$ and chose the best performance from them.}
  The experimental details are described in Appendix.~\ref{app:eeTrH}.
   
  In Fig ~\ref{res:prop:correct}, 
  we report the computational time (left) and the calculated trace (right).
  Our proposed method is significantly more efficient, and its approximation error is negligibly small.
  These benefits come from our simplification of Hessian calculation with the NNH activation function.

\subsection{Comparison with Previous Sharpness}
\label{expe:comp}

\begin{figure*}[t]
 \centering
 \begin{minipage}{73mm}
  \centering
  \includegraphics[width=\textwidth]{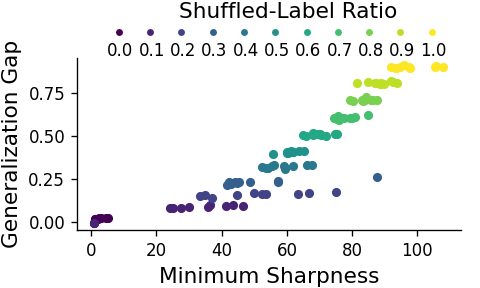}
 \end{minipage}
 \begin{minipage}{73mm}
  \centering
  \includegraphics[width=\textwidth]{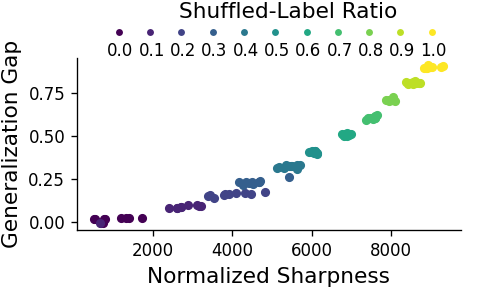}
 \end{minipage}
 \caption{Numerical experiments showing that our sharpness correlates with generalization gap. We trained an FCNNs with MNIST with partially randomized labels. The ratios are shown above.
 }
 \label{fig:exp:FCNNs}
\end{figure*}

In addition to its scale-invariance and computational efficiency, we conduct an experiment to observe how the minimum sharpness is connected to the performance of NNs.
Following previous works \cite{icml2020_3399,liang2019fisher}, we use the ``training with randomized label'' experiment to check if our sharpness has a valid correlation with the generalization gap.
In this experiment, we use corrupted MNIST as a dataset whose labels are partially shuffled with ratios $\{0.0, 0.1, ..., 1.0\}$.
We evaluate generalization gaps using two neural network architectures: fully-connected NNs (FCNNs).
We measure the difference between train accuracy and test accuracy (i.e., $\text{Acc}_{\text{train}} - \text{Acc}_{\text{test}}$) as generalization gap.
For sharpness measures, we investigate the proposed minimum sharpness and the normalized sharpness by \citet{icml2020_3399} as a baseline.\footnote{ We did not follow original procedures proposed in the work~\cite{icml2020_3399,tsuzuku2019normalized} because their approximation to diagonal elements of $\bm{H}$ is computationally heavy and numerically unstable. See more details in Appendix.~\ref{app:appDiagH} and \ref{app:eeDiagH}}
Other experimental setups are detailed in Appendix~\ref{app:exp}.

We plot the sharpness measures and the generalization gap in Fig. \ref{fig:exp:FCNNs}.
We observe that the gap and both measures are successfully correlated. 
We also carried out the same experiment using a convolutional model, LeNet, which is shown in Appendix~\ref{app:lenet}.
We claim that our minimum sharpness correlates at the same level as \citet{tsuzuku2019normalized}'s sharpness.

\section{Related Works}
\label{related_works}
\paragraph{Vulnerability of Neural Networks}
Existing studies on the robustness or defensiveness of neural networks
primarily focus on generating adversarial examples.
\citet{Szegedy2013-jn}'s pioneering work first proposed the concept of adversarial attack and found that neural network classifiers are vulnerable to some adversarial noise on input data.
Following this work, a variety of adversarial attack algorithms were developed \cite{ Goodfellow2014-ct, Moosavi-Dezfooli2016-oi, kurakin2016adversarial}.
In response to those, some works developed training algorithms to make NNs robust against such adversarial attacks on input data, i.e., adversarial training algorithms \cite{madry}.
In recent years, extending the scope of the adversarial attack to weight parameters, \citet{Sun2020-zw} showed that the robustness against weight parameters perturbations also contributes to the robustness of NNs.
Besides their novel results, it is intriguing to see that the notion they introduced, ``parameter robustness," is mathematically equivalent to a naively defined sharpness.
\paragraph{Sharpness and Generalization}
Several empirical studies have found that the sharpness of loss surfaces correlates with the generalization performance of NNs.
\citet{Hochreiter1997-xr} first observed that flat minima of shallow NNs generalize well.
Some recent results show that deep NNs also have a similar relation \cite{Keskar2016-tn, yao2018hessian}.
Further, a large-scale experiment by \citet{Jiang2019-ci} shows that sharpness measures have a stronger correlation with the generalization gap than others.
Due to its desirable property, sharpness has been implemented to some practical algorithms \cite{chaudhari2019entropy, Yi2019-rn}, and \citet{Foret2020-kh} has achieved the state-of-the-art perforce.
There are also some theoretical works rigorously formalizing the connection between sharpness, weight perturbations, and robustness of neural networks \cite{wei2019improved, tsai2021formalizing}.

\paragraph{Sharpness Measure and Scale Sensitivity}
The development of scale-invariant sharpness has emerged in response to \citet{pmlr-v70-dinh17b}'s criticism on sharpness.
\citet{wang2018identifying} utilized the rigorous PAC-Bayes theory to define a metric called pacGen, achieving negligible scale sensitivity.
\citet{liang2019fisher} utilized information geometry to design a scale-invariant measure, Fisher-Rao metric, which correlates with the generalization gap well under several scenarios.
\citet{rangamani2019scale} developed a scale-invariant sharpness measure defined over quotient manifold. 
However, their measure does not sufficiently correlate with a generalization gap.
\citet{icml2020_3399} improved \citet{wang2018identifying}'s work and achieved scale-invariant sharpness measure.
However, since the works above are exposed to either heavy computation or unrealistic approximation, they are not as suitable for practical use as the naively defined sharpness \cite{chaudhari2019entropy,Yi2019-rn,Foret2020-kh}.
Our minimum sharpness is the first scale-invariant sharpness ready for practical use.

\nocite{langley00}

\bibliography{ms}

\begin{thebibliography}{26}
\providecommand{\natexlab}[1]{#1}
\providecommand{\url}[1]{\texttt{#1}}
\expandafter\ifx\csname urlstyle\endcsname\relax
  \providecommand{\doi}[1]{doi: #1}\else
  \providecommand{\doi}{doi: \begingroup \urlstyle{rm}\Url}\fi

\bibitem[Botev et~al.(2017)Botev, Ritter, and Barber]{pmlr-v70-botev17a}
Botev, A., Ritter, H., and Barber, D.
\newblock Practical gauss-newton optimisation for deep learning.
\newblock In \emph{Proceedings of the 34th International Conference on Machine
  Learning 2017}, pp.\  557--565, 2017.

\bibitem[Chaudhari et~al.(2019)Chaudhari, Choromanska, Soatto, LeCun, Baldassi,
  Borgs, Chayes, Sagun, and Zecchina]{chaudhari2019entropy}
Chaudhari, P., Choromanska, A., Soatto, S., LeCun, Y., Baldassi, C., Borgs, C.,
  Chayes, J., Sagun, L., and Zecchina, R.
\newblock Entropy-sgd: Biasing gradient descent into wide valleys.
\newblock \emph{Journal of Statistical Mechanics: Theory and Experiment},
  2019\penalty0 (12):\penalty0 124018, 2019.

\bibitem[Dinh et~al.(2017)Dinh, Pascanu, Bengio, and Bengio]{pmlr-v70-dinh17b}
Dinh, L., Pascanu, R., Bengio, S., and Bengio, Y.
\newblock Sharp minima can generalize for deep nets.
\newblock In \emph{Proceedings of the 34th International Conference on Machine
  Learning}, pp.\  1019--1028, 2017.

\bibitem[Foret et~al.(2020)Foret, Kleiner, Mobahi, and Neyshabur]{Foret2020-kh}
Foret, P., Kleiner, A., Mobahi, H., and Neyshabur, B.
\newblock {Sharpness-Aware} minimization for efficiently improving
  generalization.
\newblock October 2020.

\bibitem[Goodfellow et~al.(2013)Goodfellow, Warde-Farley, Mirza, Courville, and
  Bengio]{pmlr-v28-goodfellow13}
Goodfellow, I., Warde-Farley, D., Mirza, M., Courville, A., and Bengio, Y.
\newblock Maxout networks.
\newblock In \emph{Proceedings of the 30th International Conference on Machine
  Learning}, pp.\  1319--1327, 2013.

\bibitem[Goodfellow et~al.(2014)Goodfellow, Shlens, and
  Szegedy]{Goodfellow2014-ct}
Goodfellow, I.~J., Shlens, J., and Szegedy, C.
\newblock Explaining and harnessing adversarial examples.
\newblock December 2014.

\bibitem[Hochreiter \& Schmidhuber(1997)Hochreiter and
  Schmidhuber]{Hochreiter1997-xr}
Hochreiter, S. and Schmidhuber, J.
\newblock Flat minima.
\newblock \emph{Neural computation}, 9:\penalty0 1--42, 02 1997.

\bibitem[Jiang et~al.(2019)Jiang, Neyshabur, Mobahi, Krishnan, and
  Bengio]{Jiang2019-ci}
Jiang, Y., Neyshabur, B., Mobahi, H., Krishnan, D., and Bengio, S.
\newblock Fantastic generalization measures and where to find them.
\newblock \emph{arXiv preprint arXiv:1912.02178}, 2019.

\bibitem[Keskar et~al.(2017)Keskar, Mudigere, Nocedal, Smelyanskiy, and
  Tang]{Keskar2016-tn}
Keskar, N.~S., Mudigere, D., Nocedal, J., Smelyanskiy, M., and Tang, P. T.~P.
\newblock On large-batch training for deep learning: Generalization gap and
  sharp minima.
\newblock \emph{arXiv preprint arXiv:1609.04836}, 2017.

\bibitem[Kurakin et~al.(2016)Kurakin, Goodfellow, Bengio,
  et~al.]{kurakin2016adversarial}
Kurakin, A., Goodfellow, I., Bengio, S., et~al.
\newblock Adversarial examples in the physical world, 2016.

\bibitem[LeCun(2019)]{lecun20191}
LeCun, Y.
\newblock 1.1 deep learning hardware: Past, present, and future.
\newblock In \emph{2019 IEEE International Solid-State Circuits
  Conference-(ISSCC)}, pp.\  12--19. IEEE, 2019.

\bibitem[Liang et~al.(2019)Liang, Poggio, Rakhlin, and Stokes]{liang2019fisher}
Liang, T., Poggio, T., Rakhlin, A., and Stokes, J.
\newblock Fisher-{R}ao metric, geometry, and complexity of neural networks.
\newblock In \emph{The 22nd International Conference on Artificial Intelligence
  and Statistics}, pp.\  888--896, 2019.

\bibitem[Madry et~al.(2018)Madry, Makelov, Schmidt, Tsipras, and Vladu]{madry}
Madry, A., Makelov, A., Schmidt, L., Tsipras, D., and Vladu, A.
\newblock Towards deep learning models resistant to adversarial attacks.
\newblock In \emph{International Conference on Learning Representations
  (ICLR)}, 2018.

\bibitem[Moosavi-Dezfooli et~al.(2016)Moosavi-Dezfooli, Fawzi, and
  Frossard]{Moosavi-Dezfooli2016-oi}
Moosavi-Dezfooli, S.-M., Fawzi, A., and Frossard, P.
\newblock Deepfool: a simple and accurate method to fool deep neural networks.
\newblock In \emph{Proceedings of the {IEEE} conference on computer vision and
  pattern recognition}, pp.\  2574--2582. cv-foundation.org, 2016.

\bibitem[Nair \& Hinton(2010)Nair and Hinton]{ICML-2010-NairH}
Nair, V. and Hinton, G.~E.
\newblock Rectified linear units improve restricted boltzmann machines.
\newblock In \emph{Proceedings of the 27th International Conference on Machine
  Learning}, pp.\  807--814, 2010.

\bibitem[Rangamani et~al.(2019)Rangamani, Nguyen, Kumar, Phan, Chin, and
  Tran]{rangamani2019scale}
Rangamani, A., Nguyen, N.~H., Kumar, A., Phan, D., Chin, S.~H., and Tran, T.~D.
\newblock A scale invariant flatness measure for deep network minima.
\newblock \emph{arXiv preprint arXiv:1902.02434}, 2019.

\bibitem[Sun et~al.(2020)Sun, Zhang, Ren, Luo, and Li]{Sun2020-zw}
Sun, X., Zhang, Z., Ren, X., Luo, R., and Li, L.
\newblock Exploring the vulnerability of deep neural networks: A study of
  parameter corruption.
\newblock June 2020.

\bibitem[Szegedy et~al.(2013)Szegedy, Zaremba, Sutskever, Bruna, Erhan,
  Goodfellow, and Fergus]{Szegedy2013-jn}
Szegedy, C., Zaremba, W., Sutskever, I., Bruna, J., Erhan, D., Goodfellow, I.,
  and Fergus, R.
\newblock Intriguing properties of neural networks.
\newblock December 2013.

\bibitem[Tsai et~al.(2021)Tsai, Hsu, Yu, and Chen]{tsai2021formalizing}
Tsai, Y.-L., Hsu, C.-Y., Yu, C.-M., and Chen, P.-Y.
\newblock Formalizing generalization and robustness of neural networks to
  weight perturbations.
\newblock \emph{arXiv preprint arXiv:2103.02200}, 2021.

\bibitem[Tsuzuku et~al.(2019)Tsuzuku, Sato, and
  Sugiyama]{tsuzuku2019normalized}
Tsuzuku, Y., Sato, I., and Sugiyama, M.
\newblock Normalized flat minima: Exploring scale invariant definition of flat
  minima for neural networks using pac-bayesian analysis.
\newblock \emph{arXiv preprint arXiv:1901.04653}, 2019.

\bibitem[Tsuzuku et~al.(2020)Tsuzuku, Sato, and Sugiyama]{icml2020_3399}
Tsuzuku, Y., Sato, I., and Sugiyama, M.
\newblock Normalized flat minima: Exploring scale invariant definition of flat
  minima for neural networks using pac-bayesian analysis.
\newblock \emph{Proceedings of the 37th International Conference on Machine
  Learning}, pp.\  6262--6273, 2020.

\bibitem[Wang et~al.(2018)Wang, Keskar, Xiong, and Socher]{wang2018identifying}
Wang, H., Keskar, N.~S., Xiong, C., and Socher, R.
\newblock Identifying generalization properties in neural networks.
\newblock \emph{arXiv preprint arXiv:1809.07402}, 2018.

\bibitem[Wei \& Ma(2019)Wei and Ma]{wei2019improved}
Wei, C. and Ma, T.
\newblock Improved sample complexities for deep neural networks and robust
  classification via an all-layer margin.
\newblock In \emph{International Conference on Learning Representations}, 2019.

\bibitem[Wu et~al.(2017)Wu, Zhu, and Weinan]{Wu2017-ox}
Wu, L., Zhu, Z., and Weinan, E.
\newblock Towards understanding generalization of deep learning: Perspective of
  loss landscapes.
\newblock June 2017.

\bibitem[Yao et~al.(2018)Yao, Gholami, Lei, Keutzer, and
  Mahoney]{yao2018hessian}
Yao, Z., Gholami, A., Lei, Q., Keutzer, K., and Mahoney, M.~W.
\newblock Hessian-based analysis of large batch training and robustness to
  adversaries.
\newblock In Bengio, S., Wallach, H., Larochelle, H., Grauman, K.,
  Cesa-Bianchi, N., and Garnett, R. (eds.), \emph{Advances in Neural
  Information Processing Systems 31}, pp.\  4949--4959, 2018.

\bibitem[Yi et~al.(2019)Yi, Zhang, Chen, Ma, and Liu]{Yi2019-rn}
Yi, M., Zhang, H., Chen, W., Ma, Z.-M., and Liu, T.-Y.
\newblock {BN-invariant} sharpness regularizes the training model to better
  generalization.
\newblock In \emph{{IJCAI}}, pp.\  4164--4170. researchgate.net, 2019.

\end{thebibliography}
\bibliographystyle{icml2021}
\appendix

\section{Proof of Proposition~\ref{prop:BP}}
\label{app:BP}
We provide proof of the proposition~\ref{prop:BP} using non-bias fully-connected neural networks (FCNNs) with NNH activation function for simplicity.
This discussion can be applied to other models including with-bias NNs and convolutional NNs.
Let $f$ be a neural network
  and 
  $\{(\mathbf{x},y)\}_i^n$ be a set of data pairs.
For a given input $(\mathbf{x},y)$,
  we denote $o_l = \langle \mathbf{w}_l, f(\mathbf{x})\rangle$ as logit for label $l$,
  softmax probability
  $p(l \mid \mathbf{x}) = p_l = \frac{1}{Z} \exp[o_l]$,
  and normalization term
  $Z=\sum_I \exp[o_l]$.
We also set $L= \frac{1}{n} \sum_i^n L_i$
  where
  $L_i = -\ln p(y_i \mid \mathbf{x}_i)$
  as loss,
  and $\Delta = \nabla^2$ as Laplace operator.

Our goal in this section is to show the following relation
\begin{align*}
  \Tr[\bm{H}] =
      \frac{1}{n}
      \sum_{i}^n
        \text{score}(\mathbf{x}_i,y_i)
\end{align*}
\begin{align*}
  \text{score}(\mathbf{x},y) 
  =
    \sum_{d}
      \left(
      \sum_l
        p_l
        \left\|\frac{\partial o_l}{\partial \bm{W}_d}\right\|_F
    -
      \left\|\frac{\partial \ln Z}{\partial \bm{W}_d}\right\|_F
      \right)
\end{align*}
where $\|\cdot\|_F$ denotes Frobenius norm.

Firstly, we can decompose 
  $
    \Tr[\bm{H}] 
    = 
    \Tr[\Delta L] 
    =
    \Tr[\Delta \sum_i L_i]
    =
    \sum_i
    \Tr[\Delta L_i]
  $
  thanks to the property of trace operation,
  $\Tr[\bm{A}+\bm{B}]= \Tr[\bm{A}]+\Tr[\bm{B}]$,
  and linearity of $\Delta$
  acting on the loss.
This decomposition has a critical role because it allows us to reduce requirements for computational resources via mini-batch calculation.

Secondly,
  we also decompose $\Tr[\Delta L_i]$ as follows:
Let $\{\bm{W}_d\}_d^D$ be a set of layer parameters in FCNNs.
Because trace operation sums up only diagonal elements,
  we have 
  $
  \Tr[\Delta L_i]
  = \sum_d \Tr[\bm{H}_d]
  $
  where $\bm{H}_d$ is diagonal block of Hessian matrix $\Delta L_i$ for $d$-th layer.
Due to this decomposition,
  all we need to calculate $\Tr[\Delta L]$
  is computation of data-wise $\Tr[\bm{H}_d]$.
From now on, we omit subscript $i$ when it has no ambiguity.

Here, we introduce important relation for NNs with NNH activation function:
\begin{align*}
  \bm{H}_d = (\bm{J}_d \bm{W}_D \bm{P} \bm{W}_D^\top \bm{J}_d^\top)
      \otimes
      (\mathbf{x}_d \mathbf{x}_d^\top)
\end{align*}
where $\bm{W}_D$ is a matrix packing $\mathbf{w}_l$,
$\bm{P}$ is a matrix $\diag[\mathbf{p}]-\mathbf{p} \mathbf{p}^\top$, where $\mathbf{p}$ is a vector containing $p(l\mid\mathbf{x})$,
$\bm{J}_d$ is a Jacobian matrix,
and $\mathbf{x}_d$ input vector for $d$-th layer.
The operation $\diag[\cdot]$ means embedding elements of a vector into diagonal position in matrix.
This equation originating from second deviation of NNH function will vanish.
The detailed explanation are available in previous work~\cite{pmlr-v70-botev17a}.

Exploiting this equation and properties of $\Tr[\cdot]$,  $\Tr[\bm{A}\otimes\bm{B}]=\Tr[\bm{A}]\Tr[\bm{B}]$
  and 
  $\Tr[\bm{A}\bm{B}]=\Tr[\bm{B}\bm{A}]$,
  we have the following results:
\begin{align*}
  \Tr[\bm{H}_d]
    &=
    \Tr[ (\bm{J}_d \bm{W}_D \bm{P} \bm{W}_D^\top \bm{J}_d^\top)
      \otimes
      (\mathbf{x}_d \mathbf{x}_d^\top)]
    &\\
    &=
    \Tr[ 
      \bm{J}_d \bm{W}_D \bm{P} \bm{W}_D^\top \bm{J}_d^\top
      ]
    \,\,\,
    \|\mathbf{x}_d\|_2^2.&
\end{align*}

Next,we reformulate the complicated term, $\Tr[\bm{J}_d \bm{W} \bm{P} \bm{W}^\top \bm{J}_d^\top]$,
  as follow:   
\begin{align*}
 & \Tr[\bm{J}_d \bm{W}_D \bm{P} \bm{W}_D^\top \bm{J}_d^\top]  &\\ 
    &= 
      \Tr[
        \bm{J}_d \bm{W}_D 
        (
            \diag[\mathbf{p}]
            -
            \mathbf{p} \mathbf{p}^\top 
        ) 
        \bm{W}_D^\top \bm{J}_d^\top
      ]
      &\\[4pt]
    &=
      \Tr[
        \bm{J}_d \bm{W}_D \diag[\mathbf{p}] \bm{W}_D^\top \bm{J}_d^\top
      ] 
      -
      \|\bm{J}_d \bm{W}_D \mathbf{p}\|_2^2
      &\\[4pt]
    &=
      \textstyle{\sum_l} \, p_l \| \bm{J}_d \mathbf{w}_l \|_2^2
      -
      \|\bm{J}_d \bm{W}_D \mathbf{p}\|_2^2&
\end{align*}

Applying this refomulation to the $\Tr[\bm{H}_d]$,
  we have the following relation
\begin{align*}
  &\Tr[\bm{H}_d] &\nonumber\\
  &= \Tr[ 
    \bm{J}_d \bm{W}_D \bm{P} \bm{W}_D^\top \bm{J}_d^\top
    ]
  \,\,\,
  \|\mathbf{x}_d\|_2^2
  &\\
  &=
    \textstyle{\sum_l} \, p_l 
        \| 
            \bm{J}_d \mathbf{w}_l 
        \|_2^2
    -
    \|
        \bm{J}_d \bm{W}_D \mathbf{p}
    \|_2^2
  \|\mathbf{x}_d\|_2^2
  &\\
  &=
    \textstyle{\sum_l} \, p_l 
      \| \bm{J}_d \mathbf{w}_l \|_2^2
      \|\mathbf{x}_d\|_2^2
     -
    \|\bm{J}_d \bm{W}_D \mathbf{p}\|_2^2
    \|\mathbf{x}_d\|_2^2
  &\\
  &=
    \textstyle{\sum_l} \, 
    p_l 
    \| (\bm{J}_d \mathbf{w}_l) \mathbf{x}_d^\top\|_F
    -
    \| (\bm{J}_d \bm{W}_D \mathbf{p}) \mathbf{x}_d^\top\|_F.&
\end{align*}
Here, we use the relation
$
  \|\mathbf{x}\|_2^2 \|\mathbf{y}\|_2^2
  =
  \Tr[
    \mathbf{x}^\top \mathbf{x}
    \mathbf{y}^\top \mathbf{y}
  ]
  =
  \Tr[
    \mathbf{x}
    \mathbf{y}^\top 
    \mathbf{y}
    \mathbf{x}^\top
  ]
  =
  \Tr[
    \mathbf{x}
    \mathbf{y}^\top 
    (
    \mathbf{x}
    \mathbf{y}^\top 
    )^\top
  ]
  = \|\mathbf{x}\mathbf{y}^\top \|_F
$
.

Finally,
  applying the back-propagation relations
  $
    \frac{\partial \ln Z}{\partial \bm{W}_d} 
    = (\bm{J}_d \bm{W}_D \mathbf{p}) \mathbf{x}_d^\top
  $
  and
  $
    \frac{\partial o_l}{\partial \bm{W}_d} 
    = (\bm{J}_d \mathbf{w}_l) \mathbf{x}_d^\top
  $
  to the terms inside the Frobenius norm,
  we have the formulation we want to state.

\section{Proof of Proposition~\ref{prop:invrselaw}}
\label{app:inverselaw}
In this section,
  we provide proof of the proposition~\ref{prop:invrselaw}.
Let $\bm{W}_d$ be a layer parameter in no-bias NNs,
$\phi$ be a NNH activation function, 
and $\alpha = \{\alpha_d\mid d = 1, \ldots, D\}$ be parameters of the $\alpha$-scale transformation.

We introduce the forward-propagation as the following repetition:
$
  \mathbf{x}_{d+1} = \phi(\bm{W}_d \mathbf{x}_d)
$.
We also denote 
  $o_l = \langle \mathbf{w}_l, \mathbf{x}_D \rangle$ as logit for label $l$
  and 
  $Z=\sum_I \exp[o_l]$.
We can describe back-propagation as follow:
\begin{align*}
  \frac{\partial \ln Z}{\partial \bm{W}_d} 
  &= 
  (\bm{J}_d \bm{W}_D \mathbf{p}) \mathbf{x}_d^\top
  &\\
  \frac{\partial o_l}{\partial \bm{W}_d} 
  &= 
  (\bm{J}_d \mathbf{w}_l) \mathbf{x}_d^\top&
\end{align*}
where 
$
  \bm{J}_d 
  = 
    \diag[\mathbf{x}_{d+1}'] \bm{W}_{d+1}^\top
    (
      \cdots 
      (\diag[\mathbf{x}_D'] \bm{W}_{D}^\top)
      \cdots 
    )
$.
The operation $\diag[\cdot]$ means embedding elements of a vector into diagonal position in matrix
and $\mathbf{x}_d'$ indicates element-wise deviation.

The $\alpha$-Scale transformation acting on $\{\bm{W}_d\}$
  modifies this back-propagation as follows
\begin{align*}
  \bm{J}_d 
  &= 
    \diag[\mathbf{x}_{d+1}'] \alpha_{d+1} \bm{W}_{d+1}^\top
    (
      \cdots 
      (\diag[\mathbf{x}_D'] \alpha_{D} \bm{W}_{D}^\top)
      \cdots 
    )&
  \\
  &= 
    \left(\prod_{k=d+1}^D \alpha_d \right)
    \,\,
    \diag[\mathbf{x}_{d+1}'] \bm{W}_{d+1}^\top
    (
      \cdots 
      (\diag[\mathbf{x}_D'] \bm{W}_{D}^\top)
      \cdots 
    ).&
\end{align*}
Exploiting the non-negative homogeneous property,
  we also have $\mathbf{x}_d$ obtained from the $\alpha$-transformed NNs as follow:
\begin{align*}
  \mathbf{x}_d
  &= 
    \phi 
    \left(
      \alpha_{d-1}
      \bm{W}_{d-1}
        \cdots
        \left(  
          \phi 
          \left( 
            \alpha_1
             \bm{W}_{1} \mathbf{x} 
          \right) 
        \cdots
        \right)
    \right)
  &\\
  &= 
    (\prod_{k=1}^{d-1} \alpha_{k})
    \phi 
    \left(
      \bm{W}_{d-1}
        \cdots
        \left(  
          \phi 
          \left( 
             \bm{W}_{1} \mathbf{x} 
          \right) 
        \cdots
        \right)
    \right).&
\end{align*}
Combining these two results,
  we have
\begin{align*}
  \frac{\partial \ln Z'}{\partial \bm{W}_d} 
  &= 
  \left(  
    \prod_{k=d+1}^D \alpha_d
    \bm{J}_d \bm{W} \mathbf{p}
  \right)
  \left( \prod_{k=1}^{d-1} \alpha_{k} \mathbf{x}_d^\top \right) &\\
  &=
  \left(\prod_{k=1,k\neq d}^{D} \alpha_{k}\right)
  (\bm{J}_d \mathbf{w}_l) \mathbf{x}_d^\top
  &\\
  &=
  \frac{1}{\alpha_d}
  (\bm{J}_d \mathbf{w}_l) \mathbf{x}_d^\top.&
\end{align*}
Here we exploit the constraint, $\prod_{d=1}^D \alpha_d=1$.
We can prove the result for $\frac{\partial o_l}{\partial \bm{W}_d}$ in the same way.

\begin{figure*}[ht]
 \centering
 \begin{minipage}{70mm}
  \centering
  \includegraphics[width=70mm]{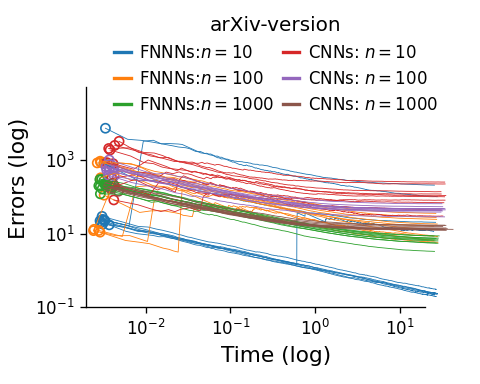}
 \end{minipage}
 \hspace{3mm}
 \begin{minipage}{70mm}
  \centering
  \includegraphics[width=70mm]{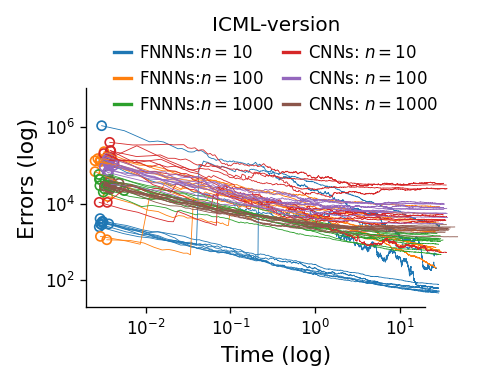}
 \end{minipage}
  \caption{
    Evaluation for approximating $\text{DIAG}[\bm{H}]$, diagonal element of Hessian matrices.
    The left and right figures are results of
      arXiv-version
      \cite{tsuzuku2019normalized}
      and 
      ICML-version
      \cite{icml2020_3399}
      respectively.
  }
  \label{fig:tsuzuku:approxi}
\end{figure*}
\begin{figure*}[t]
  \centering
  \begin{minipage}{75mm}
    \centering
    \includegraphics[width=75mm]{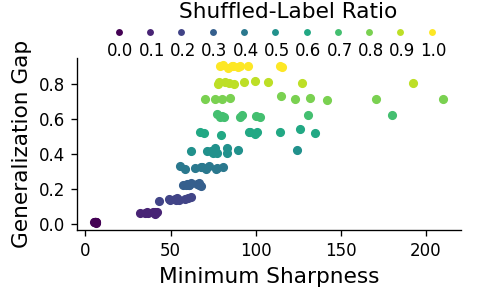}
  \end{minipage}
  \begin{minipage}{75mm}
    \centering
    \includegraphics[width=75mm]{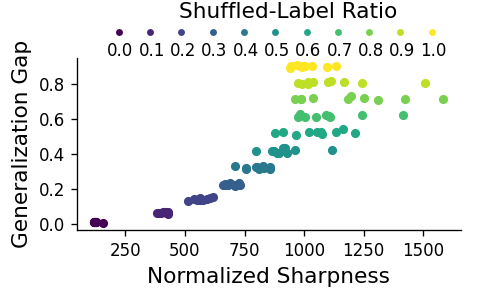}
  \end{minipage}
  \caption{Experimental results comparing minimum and normalized sharpness using LeNet. }
  \label{fig:exp:LeNet}
\end{figure*}

\section{Proof of Theorem~\ref{def:reform}}
\label{app:theorem}
In this section,
  we provide proof of the following theorem~\ref{def:reform}:
\begin{align*}
  \text{MS}_\theta
  = 
  \min_{\alpha} 
    \sum_{d=1}^D \frac{1}{\alpha_d^2} 
      \Tr[\bm{H}_{\theta,d}]
  = 
    D
    \left(
      \prod_{d=1}^D 
      \Tr[\bm{H}_{\theta,d}]
    \right)^{1/D}
\end{align*}
where $\bm{H}_{\theta,d}$ denotes diagonal blocked Hessian over $d$-th layer.

Firstly, we have the following relation enjoying proposition~\ref{prop:BP} and proposition~\ref{prop:invrselaw} acting on $\text{MS}_{\theta}$,
\begin{align*}
  &\Tr[\bm{H}_\theta]&\nonumber\\
  &=
      \frac{1}{n}
      \sum_{i=1}^n
    \sum_{d=1}^D
      \left(
      \sum_{l = 1}^K
        p_l
        \left\|\frac{\partial o_{@il}}{\partial \bm{W}_d}\right\|_F
        -
      \left\|\frac{\partial \ln Z_{@i}}{\partial \bm{W}_d}\right\|_F
      \right) 
    &\\
    &= 
    \sum_{d=1}^D
      \frac{1}{n}
      \sum_{i=1}^n
      \left(
        \sum_{l = 1}^K
        p_l
        \left\|\frac{\partial o_{@il}}{\partial \bm{W}_d}\right\|_F
        -
        \left\|\frac{\partial \ln Z_{@i}}{\partial \bm{W}_d}\right\|_F
      \right) 
    &\\
    &= 
    \sum_{d=1}^D
      \frac{1}{n}
      \sum_{i=1}^n
      \left(
      \sum_{l = 1}^K
        p_l
      \frac{1}{\alpha_d^2}
        \left\|\frac{\partial o_{@il}}{\partial \bm{W}_d}\right\|_F
    -
      \frac{1}{\alpha_d^2}
      \left\|\frac{\partial \ln Z_{@i}}{\partial \bm{W}_d}\right\|_F
      \right) 
    &\\
    &= 
    \sum_{d=1}^D
      \frac{1}{\alpha_d^2}
      \frac{1}{n}
      \sum_{i=1}^n
      \left(
      \sum_{l = 1}^K
        p_l
        \left\|\frac{\partial o_{@il}}{\partial \bm{W}_d}\right\|_F
    -
      \left\|\frac{\partial \ln Z_{@i}}{\partial \bm{W}_d}\right\|_F
      \right) 
    &\\
    &= 
    \sum_{d=1}^D
      \frac{1}{\alpha_d^2}
    \Tr[\bm{H}_{\theta,d}].&
\end{align*}
For the simplicity,
  we denote $\mathcal{H}_d=\Tr[\bm{H}_{\theta,d}]$ shortly.

We replace  $\alpha_d^{-2}$ with $\alpha_d'$.
The $\alpha_d'$ satisfies the condition of $\alpha$ in scale transformation, i.e., $\alpha_d'>0$ and $\prod_d \alpha_d' = \prod_d \alpha_d^{-2} = (\prod_d \alpha_d)^{-2} = 1$.
Hereinafter, we tackle the following minimization 
\begin{align*}
  \text{MS}_\theta
  = 
  \min_{\alpha} 
    \sum_{d=1}^D 
      \alpha_d
      \mathcal{H}_d.
\end{align*}

Thanks to the inequality of arithmetic and geometric means,
    we have the following inequality:
\begin{align*}
    \sum_{d=1}^D 
        \alpha_d
        \mathcal{H}_d
    \geq
        D
        (
        \prod_{d=1}^D 
          \alpha_d
          \mathcal{H}_d
        )^{\frac{1}{D}}
    =
        D
        \prod_{d=1}^D 
        \mathcal{H}_d^{\frac{1}{D}}
\end{align*}
where we use the relation $\prod_d \alpha_d = 1$.

The remained problem is the existence of $\alpha_d$ achieving the lower bound. 
Here, we introduce the following $\alpha$:
\begin{align*}
    \alpha_d^*
    = 
    \frac{1}{\mathcal{H}_d}
    \prod_{d=1}^D 
    \mathcal{H}_d^{\frac{1}{D}}
\end{align*}.

We can check 
achievement of the lower bound 
and 
satisfaction of the condition as $\alpha$ as follow:
\begin{align*}
    \sum_d
        \alpha_d^*
        \mathcal{H}_d
    &= 
    \sum_d
        (
        \mathcal{H}_d^{-1}
        \prod_k
        \mathcal{H}_k^{\frac{1}{D}}
        )
        \mathcal{H}_d &\\
    &=
    \sum_d
        \prod_k
        \mathcal{H}_k^{\frac{1}{D}}&\\
    &= 
        D
        \prod_d
        \mathcal{H}_d^{\frac{1}{D}},
    &
    \\
    \alpha_d^*
    &=
    \mathcal{H}_d^{-1}
    \prod_{d=1}^D 
    \mathcal{H}_d^{\frac{1}{D}}>0,
    &
    \\
    \prod_d \alpha_d^*
    &=
    \prod_d 
    (
        \mathcal{H}_d^{-1}
        \prod_k
        \mathcal{H}_k^{\frac{1}{D}}
    )&\\
    &=
    (
        \prod_d
        \mathcal{H}_d
    )^{-1}
    (
        \prod_d
        \mathcal{H}_d
    ) &\\
    &= 1. \nonumber&
    &
\end{align*}

\section{Experimental Setup for Comparison}
\label{app:exp}
  The FCNN has three FC layers with $128 $ or $512$ hidden dimensions, and LeNet has three convolution layers with $6,16$ or $120$ channels with $5$ kernel size and one FC layer with $84$ hidden dimensions.
  The optimizer is vanilla SGD 
    with $1024$ batch-size, 
    $10^{-5}$ weight decay,
    and 
    $0.9$ momentum.
  For a learning rate and the number of epochs, we set $10^{-1}$ and $3000$ for FCNNs, and $10^{-2}$ and $1000$ for LeNet.
  We use the latest parameters for NNs for evaluation.
  In the same manner, generalization gaps are calculated using the last epoch.

  To calculate the \citet{icml2020_3399}'s score,
    we exploit the exact calculation for diagonal elements of Hessian using our proposed calculation.

\section{Experiments using LeNet}
\label{app:lenet}
  In this section,
    we report the results using LeNet 
    in Fig.~\ref{fig:exp:LeNet}.
  These experiments are based on the same setting of comparison experiments
    \ref{expe:comp}.
  Even though the results of the minimum and the normalized sharpness are perfectly consistent as in the FCNN's case, both show roughly the same shape.
  Thus, we achieve the same result in the case of CNN's.





\section{Experimental Setup for Accuracy and Efficiency Calculation \texorpdfstring{$\Tr[\bm{H}]$}{Te[H]}}
\label{app:eeTrH}
  We propose an exact and efficient calculation for the trace of Hessian matrices.
  In these experiments,
    we verify the exactness
    and efficiency
    compared with
    proposed calculation 
    and baseline calculation.
  Because the baseline calculation requires heavy computation,
    we use a limited number of examples
    $\{10,100,1000\}$ in MNIST,
    a small FCNN, and a small CNN.
  The small FNNCs consists of three linear layers 
    with $20,20$ hidden dimensions.
  The small CNN has
      two convolution layers
      $20,20$ channel with $5$ kernel size
      two max-pooling layers with $2$ kernel size and $2$ stride,
      and one linear layer mapping to logits.
  We carried out experiments with $10$ different random seeds.
    
  The results are shown in Fig.\ref{res:prop:correct}.
  We compare the correct $\Tr[\bm{H}]$ obtained from baseline calculation with ones from our proposal.

  To evaluate efficiency, we compare the consumed time to calculate the values in seconds. Also, we plot the times in the same manner as the previous experiment. Note that the right figures use log-scale to show our results.
  As these results show,
    our proposal accelerates the calculation significantly.

\section{Approximation for \texorpdfstring{$\text{DIAG}[\bm{H}]$}{DIAG[H]}}
\label{app:appDiagH}

First of all,
  we define an operation $\text{DIAG}[\bm{A}]$,
  which extracts diagonal elements from a given matrix $\bm{A}$.
For simplicity,
  we roughly use this operation without paying attention to the shapes.
We note that $\diag$ and $\text{DIAG}$ are not identical operations.

   
Original normalized sharpness~\cite{icml2020_3399,tsuzuku2019normalized} ($\text{NS}$) is defined as a summation of layer-wise sharpness measures: $\text{NS} = \sum_d \text{NS}_d$
where
\begin{align*}
  \text{NS}_d
  = 
  \min_{\sigma_1, \sigma_2}
    \langle \sigma_1, \text{DIAG}[\bm{H}_d] \sigma_2 \rangle
    + 
    \langle \sigma_1^{-1}, \bm{W}_d^2 \sigma_2^{-1} \rangle.
\end{align*}
$\sigma_1, \sigma_2$ are vectors such that 
  $\prod_i \sigma_{1[i]}=1$
  and 
  $\prod_j \sigma_{2[j]}=1$, 
  similar to $\alpha$ in $\alpha$ scale transformation.
$\sigma^{-1}$ and $\bm{A}^2$ 
  indicate element-wise inverse and square operations
  respectively.

To minimize $\text{NS}$,
  we need to calculate $\text{DIAG}[\bm{H}]$.
The previous work \cite{tsuzuku2019normalized}
  approximate
  this calculation as follows
\begin{align*}
  &\text{DIAG}[\bm{H}_d]& \nonumber\\
  &\approx 
  \mathbb{E}_{\epsilon \sim \mathcal(\mathbf{0},\mathbf{1})}
  \left[
    \epsilon
    \odot
    \frac
    {
      \nabla_{\theta+r\epsilon} L(\theta+r\epsilon) 
      - 
      \nabla_{\theta-r\epsilon} L(\theta-r\epsilon) 
    }
    {2r}
  \right].&
\end{align*}
Their updated study \cite{icml2020_3399}
  proposed modified version of this approximation,
  aiming to prevent convexity of the $\text{NS}_d$ minimization 
  from violation of this approximation.

We evaluated these approximations
  with the same experiments 
  as in Appendix.~\ref{app:eeTrH}.
The results are shown in Fig.~\ref{fig:tsuzuku:approxi}.
The errors indicate 
  the L2 distance between correct diagonal Hessian $\text{DIAG}[\bm{H}]$
  and the approximation $\bm{App}$
  $\|\text{DIAG}[\bm{H}]-\bm{App}\|$.
The circles indicate the start points of each iteration.
Other settings are the same as ones used in Appendix.~\ref{app:eeTrH},
  e.g., $n$ indicates the size of data.
The left and right figures show
  that
  approximation 
  defined in 
  arXiv-version
  \cite{tsuzuku2019normalized}
  and 
  ICML-version
  \cite{icml2020_3399}
  respectively.
The results imply that 
    these approximations require
    more computational cost
    than carried in this experiment
    to be close to $\text{DIAG}[\cdot]$.

\section{Exactness and Efficiency Calculation for \texorpdfstring{$\text{DIAG}[\bm{H}]$}{DIAG[H]}}
\label{app:eeDiagH}
   
  We extend our proposed calculation of $\Tr[\bm{H}]$
    to $\text{DIAG}[\bm{HJ}]$.
  This calculation can get rid of approximations to compute normalized sharpness~\cite{icml2020_3399,tsuzuku2019normalized}.
  Due to the comparison between precision and computational cost,
    we use our calculation for normalized sharpness in our experiments.
   
  The extension is trivial
    because $\Tr[\cdot]$ and $\text{DIAG}[\cdot]$
    have similar properties.
  That is,
    $
      \Tr[\bm{A}+\bm{B}]
      =
      \Tr[\bm{A}] + \Tr[\bm{B}]
    $
    and 
    $
      \text{DIAG}[\bm{A}+\bm{B}] 
      = 
      \text{DIAG}[\bm{A}] + \text{DIAG}[\bm{B}]
    $,
    $
      \Tr[\bm{H}]
      =
      \sum_d
        \Tr[\bm{H}_d]
    $
    and 
    $
      \text{DIAG}[\bm{H}] 
      = 
      \oplus_d
        \text{DIAG}[\bm{H}_d]
    $
    where $\oplus$ denotes direct product
    ,
    and
    $\Tr[\bm{A}\otimes\bm{B}]=\Tr[\bm{A}]\Tr[\bm{B}]$
    and
    $
      \text{DIAG}[\bm{A}\otimes \bm{B}] 
      = 
      \text{DIAG}[\bm{A}] \otimes \text{DIAG}[\bm{B}]
    $.

  We can compare the performance of the approximations with our extended calculation.
  Our results in the experiment ~\ref{expe:eeTrH} are obtained by a summation of this extended $\text{DIAG}[{\bm{H}}]$ calculation: first calculate $\text{DIAG}[\bm{H}]$ and then sum $\Tr[\bm{H}]$ up.
  Per our observations, we realized accurate and significantly faster calculation.

\end{document}